\documentclass{article}
\usepackage{ijcai21}

\usepackage{times}

\usepackage{soul}
\usepackage{url}
\usepackage[hidelinks]{hyperref}
\usepackage[utf8]{inputenc}
\usepackage[small]{caption}
\usepackage{graphicx}
\usepackage{amsmath}
\usepackage{comment}
\usepackage{booktabs}
\usepackage{hyperref}
\usepackage{xcolor}
\usepackage[normalem]{ulem}

\title{Learning Football Body-Orientation as a Matter of Classification}

\author{
Adrià Arbués Sangüesa$^1$\and
Adrián Martín$^1$\and
Paulino Granero$^{2}$\and \\
Coloma Ballester$^1$\And
Gloria Haro$^1$
\\
\affiliations
$^1$Universitat Pompeu Fabra, $^2$Russian Football Union\\
\emails
\href{mailto:adria.arbues@upf.edu}{adria.arbues@upf.edu} 
}

\begin{document}

\maketitle

\begin{abstract}
Orientation is a crucial skill for football players that becomes a differential factor in a large set of events, especially the ones involving passes. However, existing orientation estimation methods, which are based on computer-vision techniques, still have a lot of room for improvement. To the best of our knowledge, this article presents the first deep learning model for estimating orientation directly from video footage. By approaching this challenge as a classification problem where classes correspond to orientation bins, and by introducing a cyclic loss function, a well-known convolutional network is refined to provide player orientation data. The model is trained by using ground-truth orientation data obtained from wearable EPTS devices, which are individually compensated with respect to the perceived orientation in the current frame. The obtained results outperform previous methods; in particular, the absolute median error is less than 12 degrees per player. An ablation study is included in order to show the potential generalization to any kind of football video footage. 
\end{abstract}

\section{Introduction}
Although deep learning (DL) has been an active field of research over the last decade, its application on top of sports data has had a slow start. The lack of universal sports datasets made it an impossible challenge for a lot of researchers, professional clubs were not aware about the unlocked potential of data-driven tools, and companies were highly focused on manual video analysis.\\
However, during this last lustrum, the whole paradigm shifted, and for instance, in the case of football, complete datasets like SoccerNet have been publicly shared
\cite{giancola2018soccernet,deliege2020soccernet}, hence providing researchers with valid resources to work with \cite{cioppa2019arthus,cioppa2020context}. At the same time, top European clubs created their own departments of data scientists while publishing their findings \cite{fernandez2018wide,llana2020right}, and companies also shifted to data-driven products based on trained large-scale models. Companies such as SciSports \cite{decroos2019actions,bransen2020player}, Sport Logiq \cite{sanford2020group}, Stats Perform \cite{sha2020end,power21} or Genius Sports \cite{quiroga2020seen} made a huge investment in research groups (in some cases, in collaboration with academia), and other companies are also sharing valuable open data \cite{metricasports,statsbomb,skillcorner}. All these facts prove that, DL is currently both trendy and useful within the context of sports analytics, thus creating a need for \textit{plug-and-play} models that could be exploited either by researchers, clubs or companies. \\
Recently, expected possession value (EPV) and expected goals models proved to produce realistic outcomes \cite{spearman2017physics,spearman2018beyond,fernandez2019decomposing}, which can be directly used by coaches to optimize their tactics. Furthermore, in this same field, Arbués-Sangüesa \textit{et al.} spotted a specific literature gap regarding the presented models: player body-orientation. The authors claimed that by merging existing methods with player orientation, the precision of existing models would improve \cite{arbues2020using}, especially in pass events. By defining orientation as the projected normal vector right in the middle of the upper-torso, the authors propose a sequential computer vision pipeline to obtain orientation data \cite{arbues2020always}. Their model stems from pose estimation, which is obtained with existing models \cite{ramakrishna2014pose,wei2016convolutional,cao2017realtime}, and achieves an absolute median error of 28 degrees, which indicates that, despite being a solid baseline, there is still room for improvement. \\ 
\begin{figure*}
\begin{center}
  \includegraphics[width=0.8\linewidth]{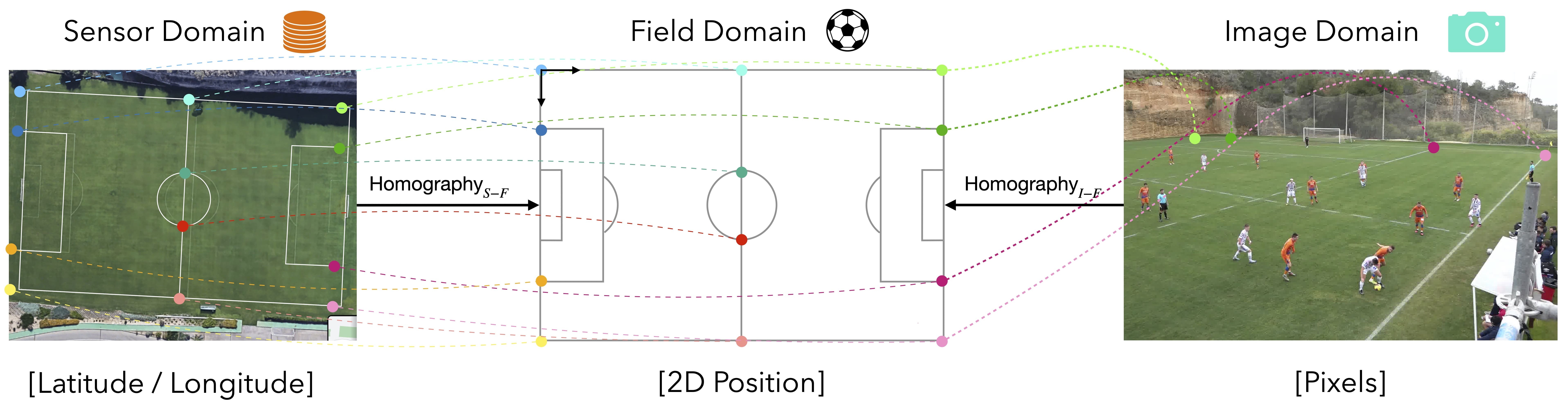}
  \caption{Several domains are merged in this research: (left) sensor-, (middle) field-, and (right) image-domain. By using corners and intersection points of field lines, the corresponding homographies are used to map data across domains into one same reference system.}
  \label{fig:ref1}
 \end{center}
\end{figure*}
Therefore, in this article, a novel deep learning 
model to obtain orientation out of any player's bounding boxes is presented. By: (1) using sensor-based orientation data as ground truth, (2) turning this estimation into a classification problem, (3) compensating angles with respect to the camera's viewing direction, and (4) introducing a cyclic loss function based on soft labels, the network is able to estimate orientation with a median error of fewer than 12 degrees per player. \\ 
The rest of the paper is organized as follows: in Section \ref{sec:data} the main data structures and types of datasets are detailed; the proposed fine-tuning process is explained in Section \ref{sec:prop} together with the appropriate details about the loss function and angle compensation. Results are shown in Section \ref{sec:res}, and finally, conclusions are drawn in Section \ref{sec:conc}. 

\section{Data Sources} \label{sec:data}
Before introducing the proposed method, a detailed description of the required materials to train this model is given. Similarly, since we are going to mix data from different sources, their corresponding domains should be listed as well: 
\begin{itemize}
    \item \textbf{Image-domain}, which includes all kinds of data related to the associated video footage. That is: (\textit{i1}) the video footage itself, (\textit{i2}) player tracking and (\textit{i3})  corners' position. Note that the result of player tracking in the image-domain consists of a set of bounding boxes, expressed in pixels; similarly, corners' location is also expressed in pixels. In this research, full HD resolution (1920 x 1080) is considered, together with a temporal resolution of 25 frames per second.
    \item \textbf{Sensor-domain}, which gathers all pieces of data generated by wearable EPTS devices. In particular, data include: (\textit{s4}) player tracking, and (\textit{s5}) orientation data. In this case, players are tracked according to the universal latitude and longitude coordinates, and orientation data are captured with a gyroscope in all XYZ Euler angles. In this work, sensor data were gathered with RealTrack Wimu wearable devices \cite{realtrack}, which generate GPS/Orientation data at 100/10 samples per second respectively. 
    \item \textbf{Field-domain}, which expresses all variables in terms of a fixed two-dimensional football field, where the top-left corner is the origin.
\end{itemize}

\begin{figure*}
\begin{center}
  \includegraphics[width=0.95\linewidth]{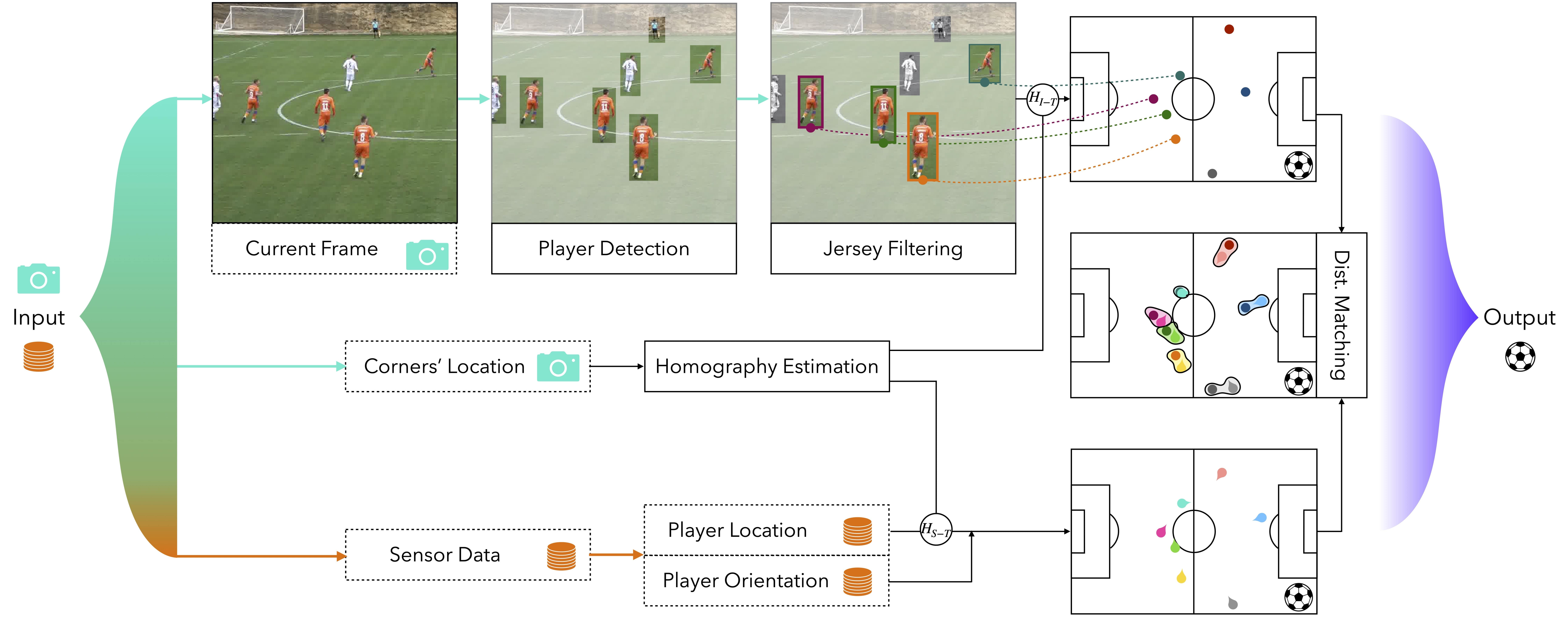}
  \caption{Proposed pipeline to match sensor orientation data with bounding boxes. Different input sources are merged: (top, image-domain) video footage, which is used for player detection and jersey filtering; the resulting bounding boxes are later mapped into the field-domain. (middle, image-domain) Corner's location, which is used for building the corresponding mapping homographies, and (bottom, sensor-domain) ground-truth data, which are also mapped into the field-domain. Finally, players in the 2D-domain are matched through pairwise distances.}
  \label{fig:pip1}
 \end{center}
\end{figure*}

Once data are gathered and synchronized from different sources, two possible scenarios are faced:
\begin{itemize}
    \item The complete case, in which all variables (\textit{i1}, \textit{i2}, \textit{i3}, \textit{s4}, \textit{s5}) are available. Note that both image- and sensor-data include unique identifiers, which are easy to match by inspecting a small subset of frames.
    \item The semi-complete case, where only part of the information is available (\textit{i1},\textit{s4},\textit{s5}). In order to estimate the missing pieces (\textit{i2, i3}) and match data across domains, a sequential pipeline is proposed in Section \ref{sec:Pip1}.
\end{itemize}
In this article, both a complete and a semi-complete datasets are used, each one containing data from single games. In particular, the complete dataset contains a full game of F.C. Barcelona's Youth team recorded with a tactical camera with almost no panning and without zoom; this dataset will be named $\text{FCB}_{DS}$. The semi-complete dataset contains a full preseason match of CSKA Moscow's professional team, recorded in a practice facility (without fans) with a single static camera that zooms quite often and has severe panning.  Similarly, this second dataset will be named $\text{CSKA}_{DS}$. Furthermore, intersection and corner image-coordinates were manually identified and labelled in more than 4000 frames of $\text{CSKA}_{DS}$ (1 frame every 37, \textit{i.e.} 1.5 seconds), with a mean of 8.3 ground-truth field-spots per frame (34000 annotations).\\

\subsection{Homography Estimation}
Since the reference system of the image- and the sensor-domain is not the same, corners' positions (or line intersections) are used to translate all coordinates into the field-domain. On the one hand, obtaining field locations in the sensor domain is pretty straightforward: since its gathered coordinates are expressed with respect to the universal latitude/longitude system, the corners' locations are fixed. By using online tools such as the \textit{Satellite View} of Google Maps, and by accurately picking field intersections, the corners' latitude and longitude coordinates are obtained. On the other hand, corner's positions in the image domain (in pixels) depend on the camera shot and change across the different frames; although several literature methods \cite{citraro2020real} can be implemented in order to get the location of these field spots or the camera pose, our proposal leverages homographies computed from manual annotations. 
From now on, the homography that maps latitude/longitude coordinates into the field will be named $H_{SF}$, whereas the one that converts pixels in the image into field coordinates will be named $H_{IF}$. The complete homography-mapping process is illustrated in Figure \ref{fig:ref1}.\\

\subsection{Automatic Dataset Completion} \label{sec:Pip1}
In this Subsection, the complete process to convert a semi-complete dataset into a complete one is described. It has to be remarked that the aim is to detect players in the image-domain and to match them with sensor data, hence pairing orientation and identified bounding boxes. Note that this procedure has been applied to $\text{CSKA}_{DS}$, which did not contain ground-truth data in the image-domain. The proposed pipeline is also displayed in Figure \ref{fig:pip1}. \\
\textbf{Player Detection}: the first step is to locate players' position in the image. In order to do so, literature detection models can be used, such as OpenPose \cite{cao2017realtime} (used in this research) or Mask R-CNN \cite{maskrcnn}. Once identified all different targets in the scene, detections are converted into bounding boxes. Note that this step does not exploit any temporal information across frames.\\ 
\textbf{Jersey Filtering}: since sensor data are only acquired for one specific team, approximately half of the detected bounding boxes (opponents) are filtered out. Given that the home/away teams of football matches are required to wear distinguishable colored jerseys, a simple clustering model can be trained. Specifically, by computing and by concatenating quantized versions of the HSV / LAB histograms, a single 48-feature vector is obtained per player. Having trained a $K$-Means model, with $K=3$, boxes with three different types of content are obtained: (1) home team, (2) away team, and (3) outliers.\\
\textbf{Mapping}: in order to establish the same reference system for both sensor and image data, all tracking coordinates are mapped into the field domain. More specifically, corner-based homographies $H_{SF}$ and $H_{IF}$ are used; in the latter, since we are dealing with bounding boxes, the only point being mapped for each box is the middle point of the bottom boundary of the box.\\
\textbf{Matching}: once all points are mapped into the field-domain, a customized version of the Hungarian method \cite{kuhn1955hungarian} is implemented, thus matching sensor and image data in terms of pairwise field-distances. \\ 

\begin{figure}
\begin{center}
  \includegraphics[width=0.8\linewidth]{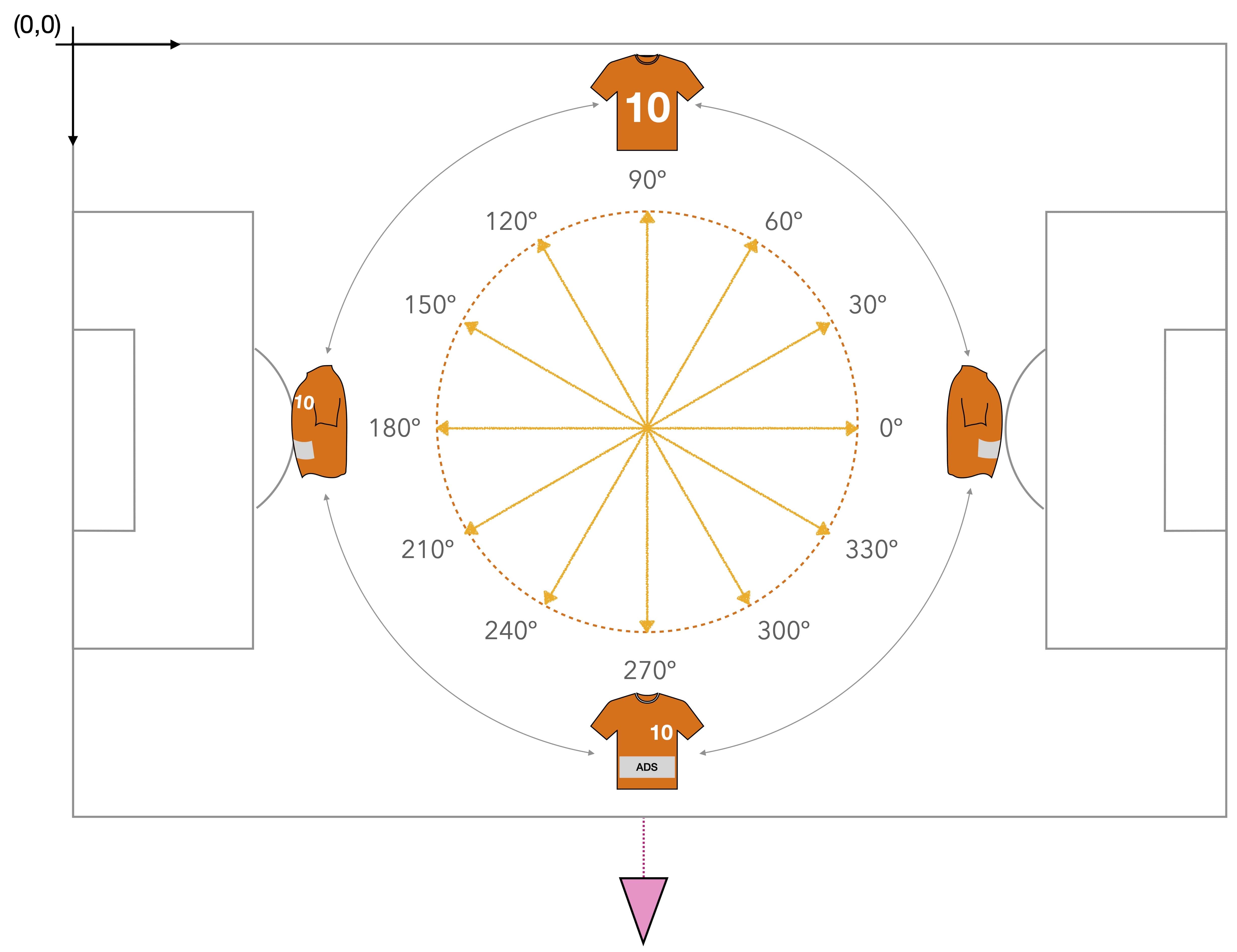}
  \caption{Orientation references in the field-domain.}
  \label{fig:ref2}
 \end{center}
\end{figure}

\begin{figure*}
\begin{center}
  \includegraphics[width=0.7\linewidth]{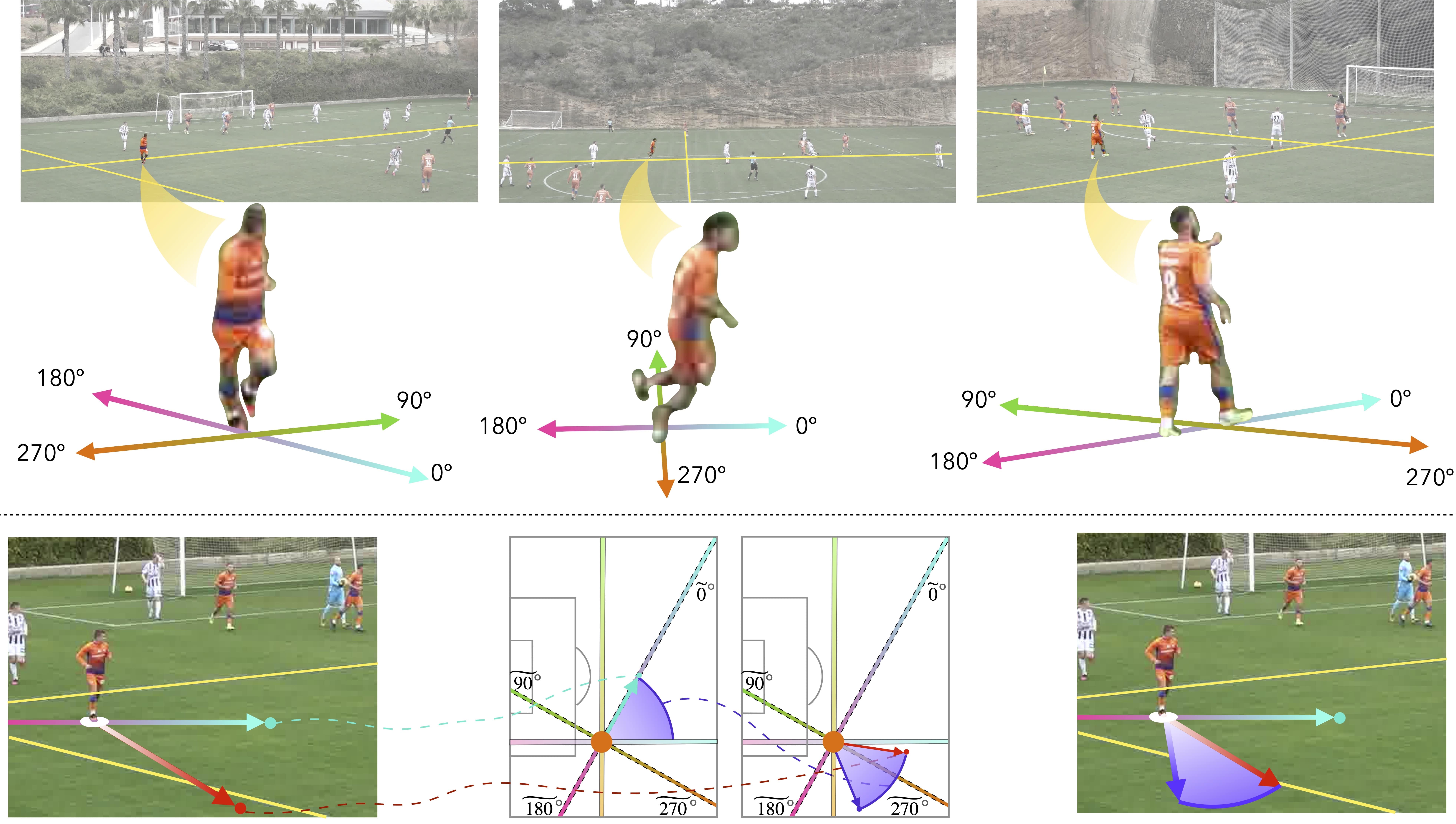}
  \caption{(Top) Three players oriented towards 0º can look really different depending on the camera pose and orientation. (Bottom) Proposed technique for angle compensation: (left) detected player together with his orientation \{red\} and \textit{apparent zero-vector} \{cyan\}; (middle-left) mapped  \textit{apparent zero-vector} in the field-domain \{dashed axes - apparent reference system, continous axes - absolute reference system\} (middle-right) Applied compensation on the original orientation \{purple\}; (right) resulting compensated absolute orientation \{purple\}. }
  \label{fig:angCompAx}
 \end{center}
\end{figure*}

\section{Proposed Method} \label{sec:prop}
In this Section, the complete adaptation and fine-tuning procedure of a state-of-the-art convolutional neural network are detailed, hence resulting in a model capable of estimating body orientation directly from bounding boxes containing players. By default, all orientations are expressed in the field-domain reference system. In this bi-dimensional field it can be assumed that 0º / 90º / 180º / 270º are the corresponding orientations of players facing towards the right / top / left / bottom sides of the fields, respectively, as shown in Figure \ref{fig:ref2}. 

\subsection{Angle Compensation}
The apparent orientation of each player is influenced by the current image content, which is drastically affected by the camera pose and its orientation. This means that, if a bounding box of a particular player is cropped without taking into account any kind of field-reference around him/her, it is not possible to obtain an absolute orientation estimation. As displayed in Figure \ref{fig:angCompAx} (top), the appearance of three players oriented towards the same direction (0 degrees) can differ a lot. Since the presented classification model only takes a bounding box as input, we propose to compensate angles \textit{a priori}, thus assuming that all orientations have been obtained under the same camera pose; \textit{i.e.} the pink camera displayed in Figure \ref{fig:ref2}. For instance, if the full chest of a player is spotted in a particular frame, its orientation must be approximately 270, no matter what the overall image context is. \\ 
In order to conduct this compensation, as seen in the bottom row of Figure \ref{fig:angCompAx}, the orientation vector of the player is first mapped into the field-domain. Then, the \textit{apparent zero-vector} is considered in the image-domain; for the reference camera, \textit{i.e.} the pink one in Figure \ref{fig:ref2}, this vector would point to the right side of the field whilst being parallel to the sidelines. By using $H_{IF}$, the \textit{apparent zero-vector} is mapped into the field-domain, and the corresponding compensation is then found by computing the angular difference between the mapped \textit{apparent zero-vector} and the reference zero-vector in the field-domain. According to Figure \ref{fig:ref2}, this difference indicates how much does the orientation vector differ from the \textit{apparent zero-vector}. \\
Formally, for a player $i$ with non-compensated orientation $\alpha_{i}'$ at position $P_{i} = (P_{i,x}, P_{i,y})$ and being the (unitary) \textit{apparent zero-vector} $Z$ described by ($1,0$), another point is defined towards the zero direction: 
\begin{equation}
    P_{i}^0 = P_{i} + Z = (P_{i,x} + 1, P_{iy})
\end{equation}
Both points $P_{i}$ and $P_{i}^0$ are mapped into the field domain by using $H_{IF}$, thus obtaining their 2D position $F_{i}$ and $F_{i}^0$, respectively.
The final compensated angle is then found as: 
\begin{equation}
    \alpha_{i} = \alpha_{i}'-\angle(\overrightarrow{F_{i}F_{i}^0}),
\end{equation}
where $\angle$ expresses the angle of the vector $\overrightarrow{F_{i}F_{i}^0}$ with respect to the reference zero-vector.

\begin{figure*}
\begin{center}
  \includegraphics[width=0.8\linewidth]{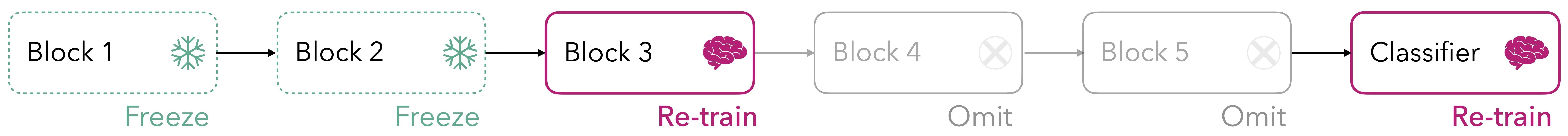}
  \caption{Proposed architecture for fine-tuning a VGG according to the main blocks of the original network.}
  \label{fig:vgg}
 \end{center}
\end{figure*}

\subsection{Network}
\begin{figure}
\begin{center}
  \includegraphics[width=0.8\linewidth]{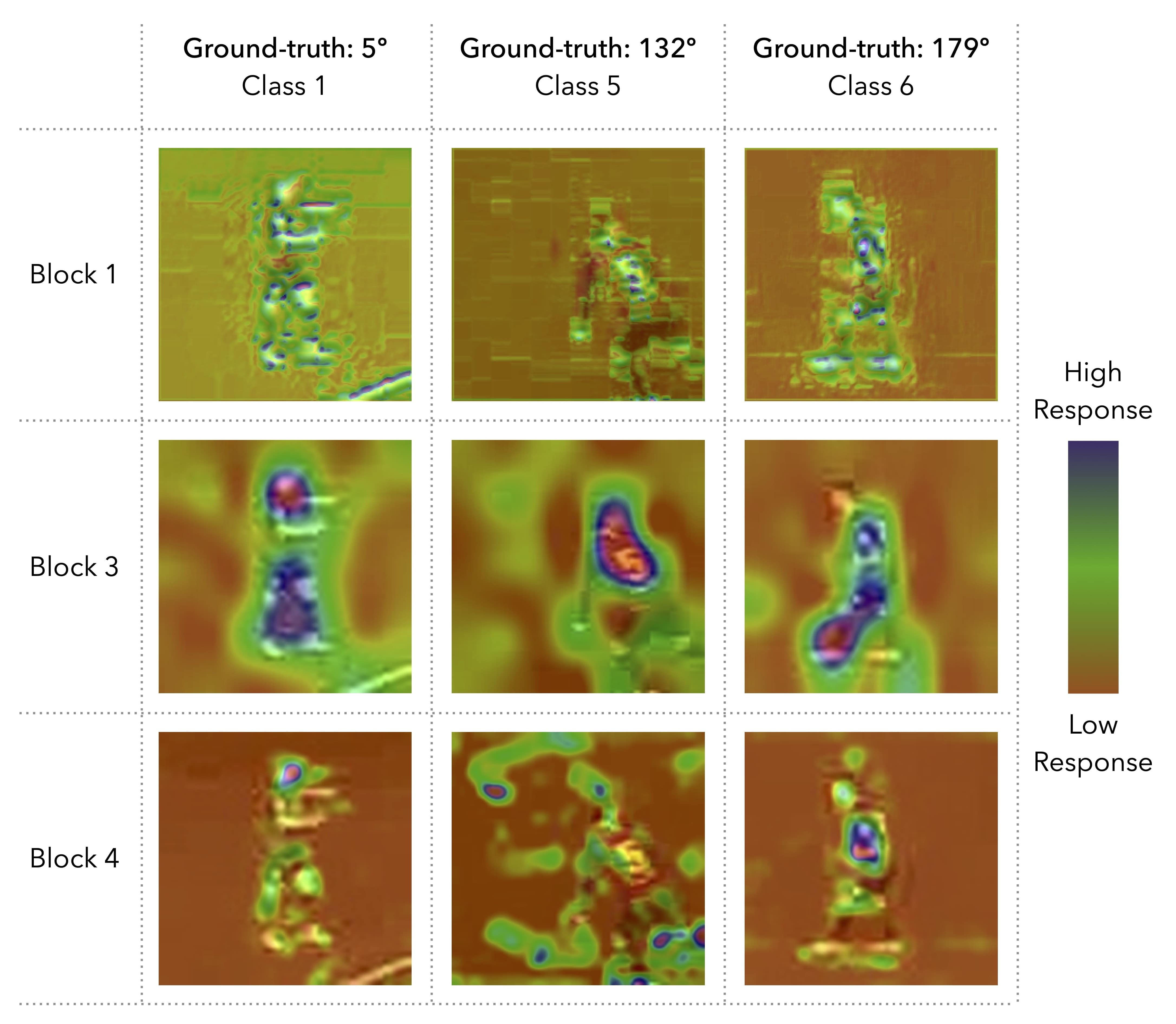}
  \caption{Obtained ScoreCam responses. While the 1st block detects mainly edges and shapes, the 3rd one has a high response over the upper-torso of players. The last row shows how the 4th block learns specific features that have little to with orientation.}
  \label{fig:weightViz}
 \end{center}
\end{figure}

Once all bounding boxes have an associated compensated body-orientation value, the model is set to be trained. In this work, orientation estimation has been approached as a classification task, where each bounding box is classified within a certain number of orientation bins. 
As detailed in Section \ref{sec:res}, orientation data are grouped into 12 bins, each one containing an orientation range of 30 degrees (\textit{e.g.} bin 1 goes from 0º to 30º, bin 2 form 30º to 60º, until bin 12, which goes from 330º to 360º). Consequently, the above-mentioned bounding boxes in the image-domain were automatically labeled with their corresponding class according to their compensated orientation. Another reason for grouping similar angles into the same class is the noisy raw orientation signals generated by the EPTS devices. \\ 
In particular, the chosen network to be fine-tuned in this research is a VGG-19 \cite{simonyan2014very}; this type of network has also been used as a backbone in existing literature methods such as OpenPose \cite{cao2017realtime}. However, in order to further analyze and to justify our choice, alternative results are shown in Section \ref{sec:res} when using Densenet \cite{huang2017densely}. The original architecture of VGG-19 is composed of 5 convolutional blocks -each one containing either 2 or 4 convolutional layers-, and a final set of fully connected layers with a probability output vector of 1000 classes. For the presented experiments, as seen in Figure \ref{fig:vgg}, the architecture adaptation and the proposed method consists of: (1) changing the dimensions of the final fully-connected layer, thus obtaining an output with a length equal to 12, the desired number of classes, (2) freezing the weights of the first couple of convolutional blocks, (3) re-training the convolutional layers of the third block and the fully connected layers of the classifier, and (4) omitting both the fourth and fifth convolutional blocks.
By visualizing the final network weights with Score-CAM \cite{wang2020score} (Figure \ref{fig:weightViz}), it can be spotted how the most important body parts regarding orientation (upper-torso) are already being vital for the sake of classification after the third block; in fact, the responses of the fourth block do not provide useful information in terms of orientation. Therefore, omitting blocks 4 and 5 is a safe choice to have an accurate model whilst decreasing the total number of parameters to be trained. \\ 
Let us finally remark that bounding boxes' values are converted into grayscale, thus improving the overall capability of generalization, since the model will not be learning the specific jersey colors, which seemed to be one of the drawbacks in \cite{arbues2020always}. In terms of data augmentation, brightnes, and contrast random changes are performed for all boxes in the training set. 

\subsection{Cyclic Loss}
An important aspect of the training process is the definition of the loss function. 
\textit{A priori}, state-of-the-art loss functions such as binary cross-entropy could be a valid resource, but in general classification scenarios, the order and the distance within classes is not taken into account. Nonetheless, in this particular scenario, we have 12 ordered-cyclic classes and a distance between them that can be well-defined. Besides, in this classification problem, since similar orientations have been grouped into bins, enforcing a one-hot encoding is not the best solution. For example, imagine a player $P_{1}$ oriented towards 31º and another $P_{2}$ oriented towards 59º; both players are included in the second bin, which encompasses all orientations between 30-60. With one-hot encoding, it would be assumed that since both $P_{1}$ and $P_{2}$ are in the second bin, both of them have the same orientation (45º). However, alternatives such as soft labels \cite{diaz2019soft} can describe the players' class as a mixture; in the given example, the soft labels of $P_{1}$/$P_{2}$ would indicate that these players are right between the first-second/second-third bins, respectively. The other challenge to be solved is the need for this loss function to be cyclic, as the first bin (number 1, 0-30º) and the last one (12, 345-360º) are actually really close. \\
Let $\{b_{1}, b_{2}, ... , b_{12}\}$ be the set of orientation classes and let $\chi = \{r_{1}, r_{2},\dots, r_{12}\}$ be the set such that each $r_j$ denotes the central angle of bin $b_{j}$, for all $j\in\{ 1,\dots,12\}$. Then, for a player $i$ with compensated ground-truth orientation $\alpha_{i}$, the soft labels representing the ground-truth probability distribution is defined as the vector with coordinates:
\begin{equation}
    y_{ij} = \frac{exp(-\phi (\alpha_{i}, r_{j}))}{\sum_{k=1}^{K}{exp(-\phi (\alpha_{i},r_{k}))}},\; \text{for}\, j=1,\dots,12
\end{equation}
where $\phi$ is cyclic distance between the ground-truth player's orientation  $\alpha_i$ and the angle corresponding to the $j$th bin, $r_{j}$:
\begin{equation}
    \phi(\alpha_{i},r_{j}) = \frac{\text{min}(|\alpha_{i}-r_{j}|,360-|\alpha_{i}-r_{j}|)^{2}}{90}.
\end{equation}
Let us denote as $x_i$ the estimated probability distribution of orientation of player $i$ obtained by applying the softmax function to the last layer of the network. Finally, our loss is the cross-entropy between $x_i$ and the ground-truth soft labels $y_i$. 


\subsection{Training Setting}
\begin{figure}
\begin{center}
  \includegraphics[width=0.65\linewidth]{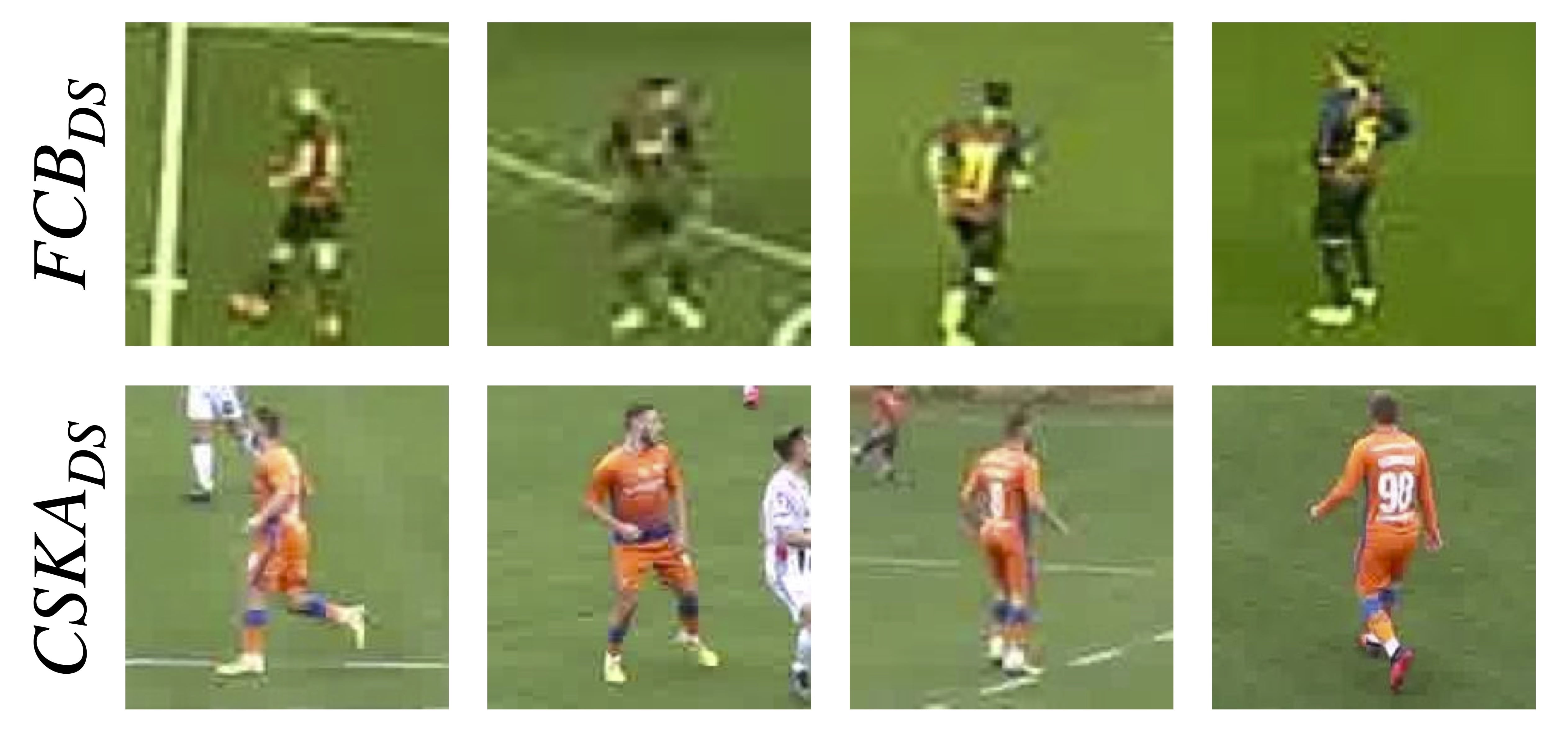}
    \caption{Resized bounding boxes of both datasets; several artifacts can be spotted in $FCB_{DS}$ (\textit{e.g.} JPEG, ringing, aliasing).}
  \label{fig:diffPl}
 \end{center}
\end{figure}

As mentioned in Section \ref{sec:data}, bounding boxes were gathered from the only two games at our disposal (Section \ref{sec:data}), both recorded under different camera shot conditions. 
Consequently, as seen in Figure \ref{fig:diffPl}, the content inside both bounding boxes differs a lot: while in $FCB_{DS}$ players are seen from a tactical camera and have small dimensions, players in $CSKA_{DS}$ are spotted from a camera that is almost in the same height as the playing field, thus resulting in big bounding boxes. Although all bounding boxes are resized as a preprocessing stage of the network, the raw datasets suffer from concept drift \cite{webb2016characterizing}.

The proposed solution in this article 
is to build an unbalanced-mixed train set; that is, merging bounding boxes from both datasets with an unbalanced distribution in the train set, whilst using the remaining instances from $FCB_{DS}$ and $CSKA_{DS}$ on their own to build the validation and the test set, respectively. In particular, the presented experiments have been carried out with a ~90-10 distribution in the training set; for each class, a total of 4500 bounding boxes -corresponding to the first half of the games- are included, where 4000 of them are obtained from $FCB_{DS}$ and the 500 remaining ones are gathered from $CSKA_{DS}$. Both the validation and test set include 500 samples per class, all of them belonging to data from the second half. 

\section{Results} \label{sec:res}
\begin{figure}
\begin{center}
  \includegraphics[width=0.75\linewidth]{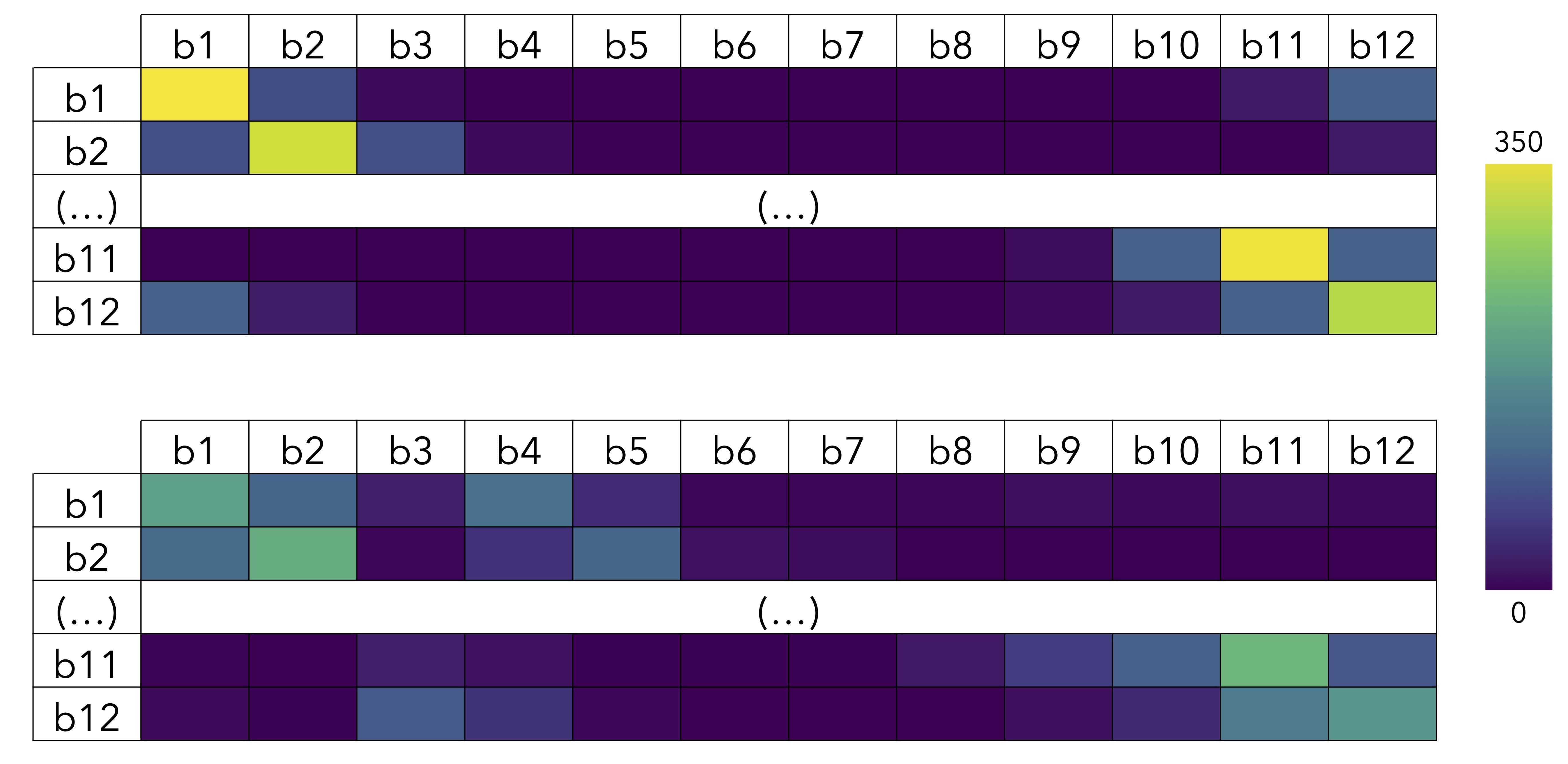}
    \caption{First and last rows of the obtained confusion matrix (test set) when using the (top) proposed cyclic and (bottom) binary cross-entropy as a loss function ($t_{12}$ and $t_{12CE}$ respectively).}
  \label{fig:cmTests}
 \end{center}
\end{figure}

The obtained classification results will be shown in terms of angular difference and confusion matrices. Nonetheless, it has to be remarked that, when grouping orientations, an intrinsic error is introduced: assuming that each bin contains $d$ degrees, and that a players' orientation is equal to the central bin value, properly classified players may have an associated absolute error up to $d/2$. 
The results of six different experiments are shown in Table \ref{tab:res}: (1) $t_{12}$ and (2) $t_{24}$ use a VGG architecture that classifies into 12 and 24 orientation bins, respectively, both trained with compensated angles; (3) $t_{12nC}$ uses the same network as in $t_{12}$ but trained without angle compensation, and (4) $t_{12den}$ uses a DenseNet architecture -fine-tuning of the fourth dense block- that performs a 12-bin classification, (5) $t_{12CE}$ uses binary cross-entropy instead of the proposed cyclic loss, and (6) $t_{12CV}$ shows the performance of the existing \textit{state-of-the-art} method \cite{arbues2020always} (12 bins as well).  
Table \ref{tab:res} contains the mean absolute error (MEAE) and the median absolute error (MDAE) of the estimated angles in each experiment. As it can be spotted, the test of 12 classes is the one providing the most reliable test results in terms of generalization; in particular, classifying orientation into 24 classes produces better results in the validation set, but seemingly the model overfits and learns specific features that do not seem to generalize properly. Moreover, the model benefits from the cyclic loss implementation, as binary cross-entropy introduces errors both in the validation and in the test set due to the unknown distance between classes and the non-cyclic angular behavior.  Actually, the obtained boost with this cyclic loss is displayed in the confusion matrices of Figure \ref{fig:cmTests}. The addition of angle compensation also proves to be vital, especially in the test set, where the corresponding video footage contained a lot of panning and zooming. Besides, the performance of DenseNet does not seem to generalize either; however, it is likely that with an exhaustive trial-error procedure of freezing weights of particular layers and performing small changes in the original DenseNet structure, this architecture should be able to generalize as well. Finally, the existing computer-vision-based method, implemented without the model in charge of the coarse corroboration, performs the worst, obtaining more than 32 degrees in both absolute errors. 

\begin{table}[]
\begin{center}
\resizebox{0.35\textwidth}{!}{%
\begin{tabular}{|c|c|c|c|c|}
\hline
            & $\text{MEAE}_{v}$ & $\text{MDAE}_{v}$ & $\text{MEAE}_{t}$ & $\text{MDAE}_{t}$ \\ \hline
$t_{12}$    & 17.37             & 9.90              & \textbf{18.92}    & \textbf{11.60}    \\ \hline
$t_{24}$    & \textbf{13.13}    & \textbf{7.70}     & 24.34             & 13.01             \\ \hline
$t_{12CE}$  & 22.34             & 17.00             & 28.98             & 23.00             \\ \hline
$t_{12nC}$  & 21.47             & 14.16             & 31.75             & 24.54             \\ \hline
$t_{12den}$ & 15.22             & 10.46             & 25.27             & 17.29             \\ \hline
$t_{CV}$  & -                 & -                 & 38.23             & 32.09             \\ \hline
\end{tabular}}
\caption{Obtained results in all experiments, expressed in terms of the mean/median absolute error, both in the validation and test set.}
\label{tab:res}
\end{center}
\end{table}

\section{Conclusions} \label{sec:conc}
In this article, a novel DL model to estimate the orientation of football players is presented. More concretely, the fine-tuned model learns how to classify players' crops into orientation bins. The core of this method combines a VGG structure with frozen and re-trained layers, an angle compensation strategy to get rid of the camera behavior, and a cyclic loss function based on soft labels that take the intra-class distance into account. The obtained results outperform (by a large margin of more than 20 degrees) the existing state-of-the-art computer-vision method with a MDAE of 11.60 degrees in the test set. Moreover, since complete datasets are difficult to gather, a sequential-based pipeline has also been proposed, which fuses data from different domains in order to establish the ground truth orientation of the player (sensor-domain) in each bounding box (image-domain). 
The main limitation of the presented model is that only two different games were used in the given dataset, as ground-truth sensor data (together with high-quality frames) are difficult to obtain. This research shows that even with unbalanced training sets it is possible to train a model with certain generalization capabilities, hence promising results should be obtained with a more varied and balanced dataset in terms of different games. As future work, apart from (a) analyzing the model extension to other football datasets and (b) testing the same approach in a regression-based fashion, its potential intra-sport generalization will be studied as well. Besides, the inclusion of fine-grained orientation into existing EPV models could lead to better performance. \\ 

\section*{Acknowledgements}
The authors acknowledge partial support by MICINN/FEDER UE project, ref. PGC2018-098625-B-I00, H2020-MSCA-RISE-2017 project, ref. 777826 NoMADS, EU H-2020 grant 952027 (project AdMiRe), and RED2018-102511-T. Besides, we also acknowledge RealTrack Systems' (in particular, Carlos Padilla's) and F.C. Barcelona's data support.


\end{document}